\documentclass[conference]{ieeeconf}
\IEEEoverridecommandlockouts

\usepackage[utf8]{inputenc}
\usepackage[T1]{fontenc}
\usepackage{amsmath,amssymb,amsfonts}
\usepackage{graphicx}
\usepackage{booktabs}
\usepackage{multirow}
\usepackage{microtype}
\usepackage{adjustbox}
\usepackage{xcolor}
\usepackage{url}
\usepackage[font=footnotesize]{caption}
\usepackage{pifont}

\usepackage{siunitx}
\usepackage[nocompress]{cite}
\usepackage{hyperref}
\hypersetup{
  hidelinks,
  pdftitle={ECoLAD: Selecting Anomaly Detectors for Automotive Deployment via Compute-Reduction Evaluation}
}
\graphicspath{{./}{./figs/}}

\makeatletter
\def\ps@IEEEtitlepagestyle{%
  \def\@oddfoot{\mycopyrightnotice}%
  \def\@evenfoot{}%
}
\def\mycopyrightnotice{%
  {\footnotesize \begin{minipage}{\textwidth}
  \centering
  \textcopyright~2026 IEEE. Personal use of this material is permitted. Permission from IEEE must be obtained for all other uses, in any current or future media, including reprinting/republishing this material for advertising or promotional purposes, creating new collective works, for resale or redistribution to servers or lists, or reuse of any copyrighted component of this work in other works.
  \end{minipage}}%
}
\makeatother

\begin{document}

\title{ECoLAD: Selecting Anomaly Detectors for Automotive Deployment via Compute-Reduction Evaluation}

\DeclareRobustCommand{\IEEEauthorrefmark}[1]{\smash{\textsuperscript{\footnotesize #1}}}

\author{%
  Kadir\mbox{-}Kaan \"Ozer$^{1,2}$,
  Ren\'{e} Ebeling$^{1}$,
  Markus Enzweiler$^{2}$%
  \thanks{$^{1}$Mercedes-Benz AG, Germany.
    \texttt{\{kadir.oezer, rene.ebeling\}@mercedes-benz.com}}%
  \thanks{$^{2}$Esslingen University of Applied Sciences, Germany.
    \texttt{markus.enzweiler@hs-esslingen.de}}%
}

\maketitle
\thispagestyle{IEEEtitlepagestyle}

\begin{abstract}
Automotive anomaly detectors are often selected from accuracy only
benchmarks on workstation class hardware, whereas in-vehicle
monitoring requires predictable scoring latency under limited CPU
parallelism. This mismatch can make methods that appear competitive
offline infeasible for deployment.

We present \emph{ECoLAD} (\textbf{E}fficiency \textbf{Co}mpute
\textbf{L}adder for \textbf{A}nomaly \textbf{D}etection), a
deployment-oriented evaluation protocol for automotive time-series
anomaly detection~(TSAD). ECoLAD defines a monotone
compute reduction ladder with explicit CPU thread caps, mechanical
integer only hyperparameter scaling, inference/full run timing
separation, and auditable run logs. Applied to proprietary
in-vehicle telemetry and two public benchmarks, it shows that
accuracy stability and deployment feasibility can diverge: some
deep detectors retain AUC-PR while losing feasible throughput,
whereas lightweight classical detectors sustain high scoring rates
with positive lift above the random baseline, providing a practical
screening procedure for detector selection under
deployment relevant constraints.
\end{abstract}

\section{Introduction}
\label{sec:intro}

Modern vehicles continuously produce telemetry from powertrain
controllers, chassis actuators, electronic control units~(ECUs),
and body systems. Detecting anomalies in these streams supports
early fault discovery, predictive maintenance, and safety
monitoring. Onboard deployment imposes hard constraints: inference
latency must be predictable, CPU parallelism is often limited to
a single thread, and memory bandwidth is restricted.

Existing time-series anomaly detection~(TSAD) benchmarks evaluate
methods under unconstrained execution on workstation hardware and
report accuracy as the sole criterion. This misrepresents deployment
feasibility in two distinct ways. First, method rankings can
change when compute budgets and CPU parallelism are jointly
reduced. Second, a method that tops an accuracy leaderboard may
become throughput infeasible on constrained hardware with no
accuracy degradation at all, an effect entirely invisible in
accuracy only rankings.

ECoLAD addresses both failure modes by specifying: (i)~a monotone
compute reduction ladder with four tiers that scale model capacity
and thread count jointly; (ii)~mechanically determined,
integer only scaling rules applied uniformly across method
families without per tier retuning; and (iii)~a throughput target
sweep reporting coverage and best achievable AUC-PR under each
target. We contribute both the \emph{protocol} (reusable on any
hardware and dataset) and an \emph{empirical instantiation} on
automotive telemetry and public benchmarks.

Prior TSAD evaluation
works~\cite{WenigEtAl2022TimeEval,SchmidlEtAl2022Anomaly,%
Zhangexperimentaleval2023,qiu2025tabunifiedbenchmarkingtime,%
si2024timeseriesbenchindustrialgradebenchmarktime,Lavin_2015}
standardize datasets, metrics, and execution environments.
ECoLAD differs in formalizing \emph{how model capacity and
parallelism are reduced} under deployment pressure, adding
explicit laddering, thread caps, and throughput feasibility
analysis as first-class protocol variables.

\paragraph{Industrial use case}
In an automotive development workflow, ECoLAD can be used before integration to discard detectors that cannot sustain the required scoring rate under CPU-thread limits, then rank the remaining candidates by detection quality. This makes the protocol a reproducible feasibility gate before calibration and integration into vehicle-side monitoring pipelines.

\begin{figure}[t]
\centering
\includegraphics[width=1.0\linewidth]{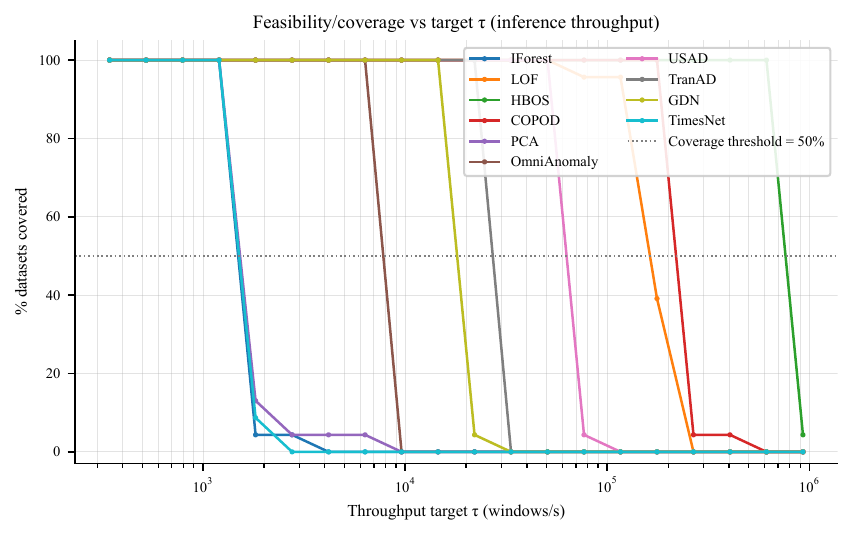}
\caption{Coverage vs.\ throughput target~$\tau$ on the
\texttt{CPU-1T} tier ($\mathrm{wps} = N/t_{\mathrm{inf}}$,
Sec.~\ref{sec:proxies}). Each curve shows the fraction of
evaluated entities whose inference throughput exceeds~$\tau$.
The horizontal line marks the 50\% feasibility reference.
Method abbreviations: IForest: Isolation Forest; LOF: Local
Outlier Factor; HBOS: Histogram-Based Outlier Score; COPOD:
Copula-Based Outlier Detection; PCA: linear subspace baseline
(see Table~\ref{tab:models_list}).}
\label{fig:budget_curve}
\end{figure}

\section{ECoLAD Protocol}
\label{sec:protocol}

\subsection{Compute Reduction Ladder}
\label{sec:tiers}

ECoLAD defines four tiers (Table~\ref{tab:tier_defs}). Each tier
fixes (i)~the execution backend, (ii)~a CPU thread cap enforced
via PyTorch \texttt{set\_num\_threads},
\texttt{set\_num\_interop\_threads}, \texttt{OMP\_NUM\_THREADS},
and BLAS limits before each run, and (iii)~a compute reduction
factor $s \in \{1.0, 0.75, 0.50, 0.25\}$. All caps and
configuration changes are logged per run. Experiments were
conducted on an Apple MacBook Pro (M3~Max, 14 performance cores,
32\,GB unified memory). For methods without GPU-accelerated
implementations (all five classical detectors evaluated
here), the \texttt{GPU} tier reduces to the reference CPU
configuration; only the thread cap and scale semantics differ.

\begin{table}[t]
\centering
\caption{ECoLAD tier definitions. \texttt{CPU-1T} is the primary
deployment stress tier.}
\label{tab:tier_defs}
\setlength{\tabcolsep}{4pt}
\renewcommand{\arraystretch}{1.05}
\scriptsize
\begin{tabular}{@{}llcc@{}}
\toprule
\textbf{Tier} & \textbf{Backend} & \textbf{Threads} & \textbf{Scale $s$} \\
\midrule
\texttt{GPU}    & GPU (M3 Max)       & uncapped & 1.00 \\
\texttt{CPU-MT} & CPU multi-thread   & 14       & 0.75 \\
\texttt{CPU-LT} & CPU limited-thread & 7        & 0.50 \\
\texttt{CPU-1T} & CPU single-thread  & 1        & 0.25 \\
\bottomrule
\end{tabular}
\end{table}

\subsection{Mechanical Hyperparameter Scaling}
\label{sec:scaling}

Each method's baseline configuration is transformed by
role-specific integer only rules. No per tier retuning is
performed. Let $s$ be the tier reduction factor:
\begin{align}
v'_{\text{work}}   &= \max(1,\,\mathrm{round}(s\,v)),         \label{eq:work}   \\
v'_{\text{width}}  &= \max(1,\,\mathrm{round}(\sqrt{s}\,v)),  \label{eq:width}  \\
v'_{\text{heads}}  &= \max(1,\,\mathrm{round}(\sqrt{s}\,v)),  \label{eq:heads}  \\
v'_{\text{depth}}  &= \max(1,\,\mathrm{round}(s^{1/4} v)),    \label{eq:depth}  \\
v'_{\text{window}} &= \max(8,\,\mathrm{round}(\sqrt{s}\,v)).  \label{eq:window}
\end{align}
Width and heads scale with $\sqrt{s}$ because model capacity
scales quadratically with width, so $\sqrt{s}$ achieves a
roughly proportional capacity reduction. Depth scales more
slowly ($s^{1/4}$) to prevent collapse of shallow models.
Parameters governing decision semantics
(e.g., contamination thresholds) are not scaled.

\paragraph{Scaling fairness across method families}
The five classical detectors (Table~\ref{tab:models_list}) carry
no architectural width, depth, or attention-head parameters; only
the work rule~\eqref{eq:work} applies (e.g., number of trees or
histogram bins). Deep models lose width, depth, and heads
simultaneously, producing a larger capacity reduction at low~$s$.
This asymmetry reflects genuine architectural differences, not a
protocol defect. Constraint repairs were required in fewer than
4\% of runs, affecting only TranAD and GDN at \texttt{CPU-LT}
and \texttt{CPU-1T}.

\subsection{Throughput and Feasibility}
\label{sec:proxies}

Inference only time $t_{\mathrm{inf}}$ is preferred over full run
time $t_{\mathrm{e2e}}$ to avoid conflating offline training cost
with online scoring capacity:
\begin{equation}
t_{\mathrm{inf}} :=
\begin{cases}
\tilde{t}_{\mathrm{inf}}, & \text{if scoring only time is instrumented,} \\
t_{\mathrm{e2e}},         & \text{otherwise.}
\end{cases}
\label{eq:tinf}
\end{equation}
The timing source is logged per run. Throughput is
$\mathrm{wps} = N/t_{\mathrm{inf}}$ (windows per second), where
$N = T - w + 1$ for windowed methods and $N = T$ for
nonwindowed methods. An entity is \emph{feasible} at target
$\tau$ if $\mathrm{wps} \ge \tau$. The reference point
$\tau = 500$~wps corresponds to scoring at 500\,Hz (2\,ms per
sample) under unit stride windowing; if $\tau$ is unmet, scoring
latency grows without bound under sustained streaming.

For each (method, dataset, $\tau$), achievable AUC-PR is the
best value among measured configurations satisfying
$\mathrm{wps} \ge \tau$. Coverage is the fraction of entities
for which at least one configuration qualifies.

\section{Experimental Setup}
\label{sec:setup}

We evaluate ten detectors spanning classical, deep,
attention-based, and graph-based families
(Table~\ref{tab:models_list}). Classical baselines are
implemented via PyOD~\cite{zhao_pyod_2019}; deep methods follow
their original training and scoring procedures. All methods are
trained in an unsupervised or self-supervised fashion; labels are
used only for evaluation.

\begin{table}[t]
\centering
\caption{Evaluated detectors.}
\label{tab:models_list}
\setlength{\tabcolsep}{4.2pt}
\renewcommand{\arraystretch}{1.05}
\scriptsize
\begin{tabular}{@{}p{0.48\linewidth}p{0.44\linewidth}@{}}
\toprule
\textbf{Method} & \textbf{Reference} \\
\midrule
IForest (Isolation Forest)         & \cite{liu_isolation_2008} \\
LOF (Local Outlier Factor)         & \cite{breunig_lof_2000} \\
HBOS (Histogram-Based Outlier Score) & \cite{goldstein_hbos_2012} \\
COPOD (Copula-Based Outlier Det.)  & \cite{li_copod_2020} \\
PCA baseline                       & (linear subspace) \\
\midrule
USAD                               & \cite{audibert_usad_2020} \\
TranAD                             & \cite{tuli_tranad_2022} \\
OmniAnomaly                        & \cite{su_robust_2019} \\
GDN (Graph Deviation Network)           & \cite{deng_graph_2021} \\
TimesNet                           & \cite{wu_timesnet_2023} \\
\bottomrule
\end{tabular}
\end{table}

\noindent\textbf{Datasets.}
\emph{Telemetry}~(proprietary): a single in-vehicle multivariate
recording with 80{,}000 datapoints and 19 synchronized
powertrain/chassis/motion features; 40{,}000 training /
40{,}000 test points; labeled anomaly rate~0.022 (random scorer
AUC-PR baseline~0.022). \emph{SMD}~(Server Machine
Dataset~\cite{su_robust_2019}): 22 multivariate
server-monitoring entities; used for RQ2 degradation analysis
to verify that patterns generalize beyond vehicle telemetry.
\emph{SMAP}~(Soil Moisture Active Passive~\cite{Hundman_2018}):
multivariate spacecraft telemetry included in RQ3 to avoid
overfitting feasibility conclusions to a single dataset. The
Telemetry results should be interpreted as an industrial case
study rather than a statistically exhaustive fleet-level
validation. All results are averaged over two random seeds;
seed to seed AUC-PR standard deviations for neural methods
range from 0.000 to 0.003.

Three research questions guide the analysis: (RQ1)~how AUC-PR
and ranking change across compute tiers; (RQ2)~which degradation
mode dominates (throughput bottleneck or accuracy drift);
(RQ3)~how coverage and achievable AUC-PR change under
throughput targets~$\tau$ at \texttt{CPU-1T}.

\section{Results}
\label{sec:results}

\subsection{RQ1: Cross-Tier Detection Quality}

Table~\ref{tab:auc_all_tiers} summarizes AUC-PR across all tiers
for SMD and Telemetry. AUC-PR drift exists but is strongly
method and domain dependent.

\paragraph{SMD}
OmniAnomaly is stable near~0.51 AUC-PR and PCA is essentially
constant~(0.448) across all tiers. LOF degrades markedly
($0.145 \to 0.073$ at \texttt{CPU-1T}). Several neural baselines
(GDN, TimesNet) exhibit modest drift that can change relative
ordering even when absolute changes are small.

\paragraph{Telemetry}
The random scorer baseline is~0.022 (anomaly rate
$\pi = 0.02184$). HBOS achieves the highest AUC-PR~(${\approx}2.9\times$
above random) with only minor tier to tier drift. Several deep
methods (USAD, TranAD, OmniAnomaly) cluster near $1.9\times$
above random with minimal drift, indicating limited separability
on this signal rather than instability under compute reduction.
Top methods on SMD differ from those on Telemetry, and
tier sensitive methods such as LOF shift substantially under
constrained execution. Both effects are invisible in a
single tier leaderboard.

\begin{table}[t]
\centering
\caption{AUC-PR across compute tiers for SMD and Telemetry
(means). The Telemetry random scorer baseline is AUC-PR\,$=$\,0.022;
SMD anomaly rates vary by machine.}
\label{tab:auc_all_tiers}
\scriptsize
\setlength{\tabcolsep}{2.8pt}
\renewcommand{\arraystretch}{1.05}
\begin{adjustbox}{max width=\linewidth}
\begin{tabular}{@{}l
  *{4}{S[round-precision=3,table-format=1.3]}
  *{4}{S[round-precision=3,table-format=1.3]}
  @{}}
\toprule
\multirow{2}{*}{\textbf{Model}} &
\multicolumn{4}{c}{\textbf{SMD AUC-PR}} &
\multicolumn{4}{c}{\textbf{Telemetry AUC-PR}} \\
\cmidrule(lr){2-5}\cmidrule(lr){6-9}
& {\textbf{GPU}} & {\textbf{MT}} & {\textbf{LT}} & {\textbf{1T}}
& {\textbf{GPU}} & {\textbf{MT}} & {\textbf{LT}} & {\textbf{1T}} \\
\midrule
COPOD       & 0.250 & 0.250 & 0.250 & 0.250 & 0.035 & 0.035 & 0.035 & 0.035 \\
GDN         & 0.296 & 0.302 & 0.272 & 0.307 & 0.051 & 0.048 & 0.050 & 0.049 \\
HBOS        & 0.303 & 0.303 & 0.302 & 0.298 & 0.064 & 0.062 & 0.062 & 0.055 \\
IForest     & 0.318 & 0.317 & 0.302 & 0.274 & 0.041 & 0.042 & 0.041 & 0.043 \\
LOF         & 0.145 & 0.091 & 0.120 & 0.073 & 0.055 & 0.050 & 0.052 & 0.049 \\
OmniAnomaly & 0.504 & 0.511 & 0.512 & 0.511 & 0.041 & 0.041 & 0.041 & 0.041 \\
PCA         & 0.448 & 0.448 & 0.448 & 0.448 & 0.037 & 0.037 & 0.037 & 0.037 \\
TimesNet    & 0.280 & 0.292 & 0.282 & 0.312 & 0.057 & 0.051 & 0.055 & 0.050 \\
TranAD      & 0.360 & 0.363 & 0.359 & 0.363 & 0.041 & 0.041 & 0.041 & 0.041 \\
USAD        & 0.469 & 0.480 & 0.476 & 0.483 & 0.041 & 0.041 & 0.041 & 0.041 \\
\bottomrule
\end{tabular}
\end{adjustbox}
\end{table}

\subsection{RQ2: Degradation Modes}
\label{sec:rq2}

\begin{figure*}[t]
\centering
\includegraphics[width=1.0\linewidth]{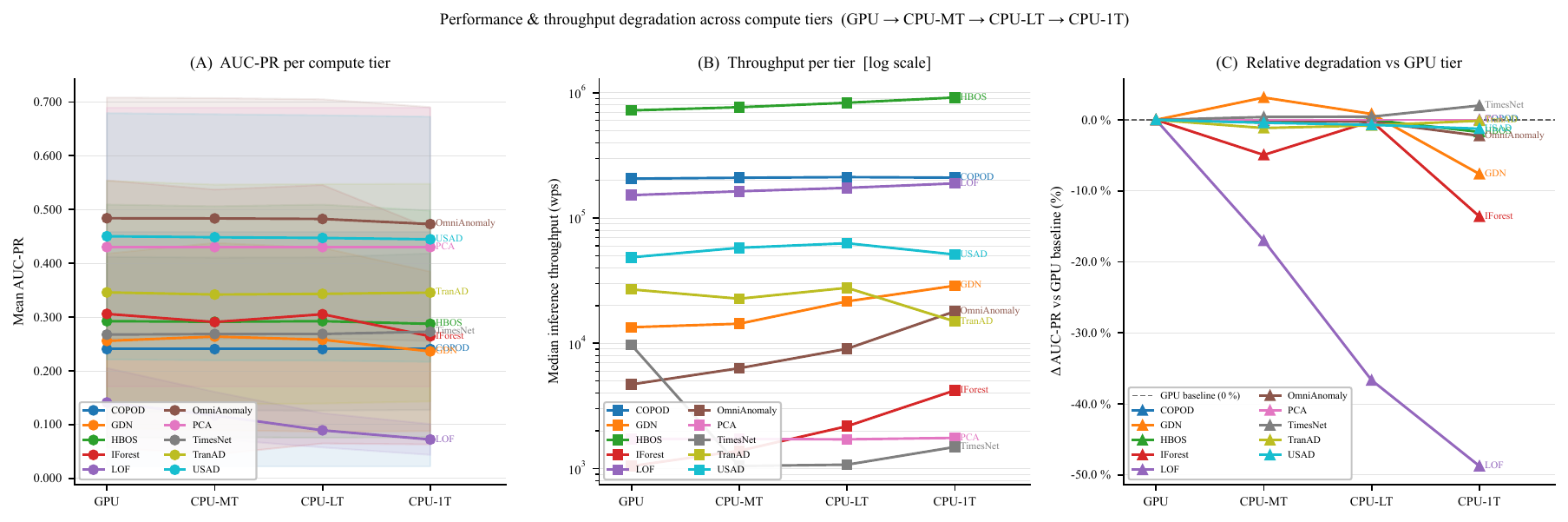}
\caption{Performance and throughput across ECoLAD compute tiers
(GPU $\to$ CPU-MT $\to$ CPU-LT $\to$ CPU-1T).
(A)~Mean AUC-PR per method.
(B)~Median inference throughput (wps, log scale;
$t_{\mathrm{inf}}$ as defined in Sec.~\ref{sec:proxies}).
(C)~Relative AUC-PR change versus the GPU tier~(\%).}
\label{fig:degradation}
\end{figure*}

Fig.~\ref{fig:degradation} and Table~\ref{tab:throughput_per_tier}
reveal three distinct degradation modes.

\noindent\textbf{Backend limited.}
TimesNet's AUC-PR is nearly flat across tiers, yet its inference
throughput on SMD drops from 9{,}569 to~1{,}483~wps at
\texttt{CPU-1T}, and on Telemetry from 11{,}164 to 1{,}751~wps.
Feasibility loss is purely throughput driven and is invisible in
accuracy only reports.

\noindent\textbf{Quality drift limited.}
LOF sustains over 76{,}000~wps on Telemetry and 193{,}000~wps
(median) on SMD at \texttt{CPU-1T}, yet shows the largest
negative $\Delta$AUC-PR under capacity reduction
(Fig.~\ref{fig:degradation}C). Runtime is not the bottleneck.

\noindent\textbf{Graceful degraders.}
HBOS and COPOD retain high throughput and near-flat AUC-PR at
all tiers. For HBOS, the reduced work scale ($s{=}0.25$) actually
\emph{increases} throughput on Telemetry from 70{,}503 to over
2{,}000{,}000~wps by reducing histogram bins per scoring call.

\begin{table}[t]
\centering
\caption{\texttt{CPU-1T} deployment snapshot: inference throughput
and full run throughput (wps). SMD: 22-entity median with p10 for
tail latency spread. Telemetry: single entity. Classical methods
show no inference/full run gap; OmniAnomaly's gap
(${\approx}23\times$ SMD, ${\approx}55\times$ Telemetry) reflects
per entity model fitting.}
\label{tab:throughput_per_tier}
\scriptsize
\setlength{\tabcolsep}{3.0pt}
\renewcommand{\arraystretch}{1.05}
\begin{adjustbox}{max width=\linewidth}
\begin{tabular}{@{}l
  S[table-format=7.0] S[table-format=7.0] S[table-format=7.0]
  @{\hspace{4pt}}
  S[table-format=7.0] S[table-format=7.0]@{}}
\toprule
& \multicolumn{3}{c}{\textbf{SMD CPU-1T}}
& \multicolumn{2}{c}{\textbf{Telemetry CPU-1T}} \\
\cmidrule(lr){2-4}\cmidrule(lr){5-6}
\textbf{Model}
  & {\textbf{infer.}} & {\textbf{p10}} & {\textbf{full run}}
  & {\textbf{infer.}} & {\textbf{full run}} \\
\midrule
COPOD       & 209288 & 204604 & 209288 & 485549 & 485549 \\
GDN         &  28693 &  28348 &   7527 &  50583 &   2955 \\
HBOS        & 913376 & 884285 & 913376 &2058696 &2058696 \\
IForest     &   4199 &   4133 &   4199 &   8015 &   8015 \\
LOF         & 193397 & 171927 & 193397 &  76420 &  76420 \\
OmniAnomaly &  18000 &  17871 &    780 &  19148 &    347 \\
PCA         &   1752 &   1658 &   1752 &   7516 &   7516 \\
TimesNet    &   1483 &   1471 &   1433 &   1751 &   1274 \\
TranAD      &  14973 &  14871 &   5890 &  27333 &   4051 \\
USAD        &  51091 &  50780 &  24455 &  85785 &  13427 \\
\bottomrule
\end{tabular}
\end{adjustbox}
\end{table}

\subsection{RQ3: Throughput-Constrained Behavior}
\label{sec:rq3}

\begin{figure*}[t]
\centering
\includegraphics[width=1.0\linewidth]{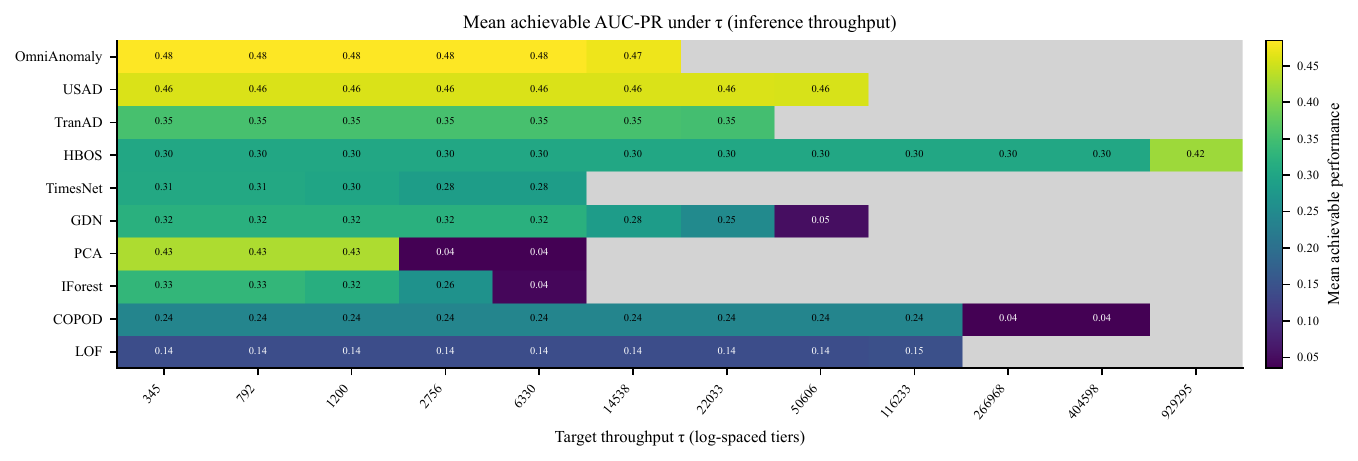}
\caption{Mean achievable AUC-PR under throughput targets on the
\texttt{CPU-1T} tier (inference throughput,
Sec.~\ref{sec:proxies}). Columns are log spaced throughput
targets~$\tau$; rows are methods. Hatched cells indicate targets
where coverage falls below 50\% of entities.}
\label{fig:heatmap_tau}
\end{figure*}

Fig.~\ref{fig:budget_curve} shows that classical detectors retain
high coverage over a wide $\tau$ range, while deep models become
infeasible at higher targets. IForest~(4{,}199~wps), PCA
(1{,}752~wps), and TimesNet~(1{,}483~wps on SMD) exhaust feasible
configurations quickly as $\tau$ rises.
Fig.~\ref{fig:heatmap_tau} shows that HBOS sustains positive
AUC-PR even at the highest feasible~$\tau$, while methods that
lose coverage early provide no operating point above the random
baseline at high throughput targets.

\section{Discussion and Limitations}
\label{sec:discussion}

ECoLAD makes compute reduction, thread caps, and throughput
feasibility explicit and auditable as first-class protocol
variables. Rank drift under constrained execution is more often
driven by architectural throughput bottlenecks than by accuracy
degradation, motivating a feasibility first filter:
screen detectors for deployment relevant scoring rates before
metric based selection. Backend sensitive methods (TimesNet) and
graceful degraders (HBOS, COPOD) are only distinguishable when
throughput is disaggregated by tier and dataset; pooled reporting
obscures both effects.

\paragraph{Platform generalizability}
The absolute throughput numbers are specific to the M3~Max and
require a platform-specific correction factor before mapping to
ECU-class targets. The qualitative degradation modes are
architecture driven, but their quantitative transfer to ECU-class hardware requires validation on the target platform.
Tail latency variability is captured by the p10/p90
inference throughput columns in
Table~\ref{tab:throughput_per_tier}; preprocessing overhead is
included in $t_{\mathrm{e2e}}$ and isolated from
$t_{\mathrm{inf}}$ for neural methods. Validation on ECU-class
hardware remains as future work.

\paragraph{Protocol scope and reproducibility}
The telemetry dataset and pipeline code are proprietary. The
protocol is specified at sufficient detail for independent
re-implementation; mechanical scaling provides a reproducible
no retuning baseline.

\section{Conclusion}
\label{sec:conclusion}

ECoLAD provides a deployment-oriented evaluation protocol for
automotive time-series anomaly detection that makes compute
reduction, CPU parallelism caps, throughput feasibility, and
auditability explicit. Across automotive telemetry and public
benchmarks, accuracy rankings shift under constrained execution
and throughput feasible operating points can exclude otherwise
competitive methods. Per tier, per dataset reporting exposes
backend sensitivity and compute reduction effects invisible in
pooled, accuracy only leaderboards. The protocol specification
is hardware and dataset agnostic; practitioners can instantiate
it on their own platform to obtain deployment relevant rankings
before committing to a detector.

\bibliographystyle{IEEEtran}
\bibliography{references}

\end{document}